\documentclass[11pt,conference]{IEEEtran}
\usepackage[T1]{fontenc}
\usepackage{amsmath,amssymb,bm}
\usepackage{graphicx,booktabs,cite,algorithm,algorithmic,placeins}
\usepackage[hidelinks]{hyperref}
\newcommand{\vect}[1]{\bm{#1}}
\newcommand{\mat}[1]{\bm{#1}}
\title{Trajectory Inversion for a Rudderless Dual Vectored-Thrust Flying Wing}
\author{\IEEEauthorblockN{Weicheng Di\IEEEauthorrefmark{1}\IEEEauthorrefmark{2}, Qiqi Liu\IEEEauthorrefmark{3}, Jinwu Xiang\IEEEauthorrefmark{2}, Daochun Li\IEEEauthorrefmark{2}, and Zhan Tu\IEEEauthorrefmark{3}}
\IEEEauthorblockA{\IEEEauthorrefmark{1}Faculty of Innovation Engineering, Macau University of Science and Technology, Macau, China}
\IEEEauthorblockA{\IEEEauthorrefmark{2}School of Aeronautical Science and Engineering, Beihang University, Beijing, China}
\IEEEauthorblockA{\IEEEauthorrefmark{3}Institute of Unmanned System, Beihang University, Beijing, China}
\IEEEauthorblockA{Corresponding author: zhantu@buaa.edu.cn}
\thanks{Weicheng Di and Qiqi Liu contributed equally to this work.}}
\IEEEoverridecommandlockouts
\begin{document}
\maketitle

\begin{abstract}
A rudderless dual vectored-thrust flying wing combines vertical takeoff and landing with efficient forward flight in a compact tailless configuration. Without aerodynamic control surfaces, vectored thrust couples translation and pitch control. Therefore, generating dynamically compatible trajectories remains an open challenge. To address this challenge, this study proposes a trajectory generation method that jointly optimizes path geometry and timing. Trajectory inversion yields state and control histories that satisfy the coupled equations of motion. The optimization balances curvature-induced acceleration against aerodynamic support, lowering thrust-vectoring demand while retaining the defining maneuver features. Representative looping and rolling sequences are tracked in closed-loop simulations. These results provide a basis for planning autonomous aerobatic maneuvers with rudderless vectored-thrust aircraft.
\end{abstract}

\section{Introduction}
Unmanned aircraft that combine vertical takeoff and landing (VTOL) with efficient forward flight can operate from confined sites while retaining the range of a fixed-wing platform~\cite{ducard2021review}. A rudderless dual vectored-thrust flying wing provides this capability with a compact tailless planform. Two propulsors control its position and attitude through rotor-speed commands and thrust-vectoring angles. Removing conventional control surfaces simplifies the structure while preserving the aerodynamic efficiency of the flying-wing layout. Hover-to-forward transition has been demonstrated on this platform~\cite{Di2025TaillessVTOL}.

Aerobatic maneuvers impose coordinated requirements on the flight path and attitude. A straight roll maintains its path while the aircraft rotates about its longitudinal axis, whereas a looping maneuver requires an ordered change in both flight direction and attitude. For this aircraft, the thrust component that produces the pitching moment also contributes to translation. Consequently, the attitude required along a path depends on acceleration and aerodynamic loads. Generating a smooth geometric curve alone leaves this coupling unresolved.

Smooth polynomial and B-spline representations provide adjustable flight paths~\cite{richter2016polynomial,usenko2017bspline}. Differential flatness has supported aerobatic trajectory generation for VTOL fixed-wing aircraft and tail-sitters under the adopted modeling assumptions~\cite{tal2022aerobatic,lu2025autonomous}. Direct numerical optimization can impose the equations of motion together with input constraints~\cite{foehn2021timeoptimal}, while stable inversion recovers internal motion left undetermined by prescribed outputs~\cite{druecker2023tracking}. For the rudderless flying wing, normal thrust participates in both force and moment balance. Its effect on internal attitude motion must therefore enter trajectory generation, particularly during large-angle maneuvers.

This study uses trajectory inversion to recover compatible attitude and propulsor-command histories, and uses these demands to guide joint optimization of path geometry and timing. The approach addresses two related problems: obtaining state and input references consistent with the force--moment coupling, and reducing acceleration and vectoring demand while preserving maneuver features. Three representative sequences illustrate the resulting looping and rolling motions in nonlinear model predictive control (NMPC)--incremental nonlinear dynamic inversion (INDI) simulations.

The rest of this paper is organized as follows. Section~\ref{sec:problem} describes the aircraft and coupling. Section~\ref{sec:method} outlines trajectory generation and tracking. Section~\ref{sec:results} presents the representative results, followed by conclusions in Section~\ref{sec:conclusion}.

\section{Aircraft and Trajectory Requirements}\label{sec:problem}
\begin{figure}[t]
\centering
\includegraphics[width=0.94\linewidth]{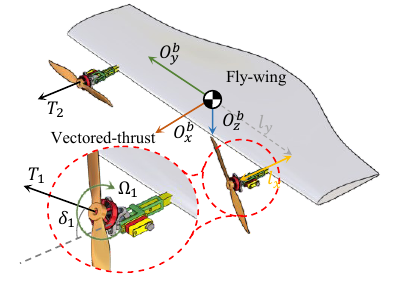}
\caption{Aircraft configuration, body axes and dual vectored-thrust actuation.}
\label{fig:configuration}
\end{figure}

The aircraft in Fig.~\ref{fig:configuration} has a mass of 0.405~kg and a wing span of 0.8~m. Body axes point forward, right and downward. Inertial position is expressed in north--east--down coordinates, while the trajectory figures display upward height. The two propulsors are displaced from the center of mass along the body longitudinal and lateral axes. Each propulsor varies its thrust magnitude and tilts its thrust vector in the body longitudinal--normal plane.

Let $\vect p$ denote inertial position, $\mat R$ the body-to-inertial rotation, $\vect\omega$ body angular velocity, and $\mat I$ the inertia matrix. The translational and rotational balances are
\begin{equation}
\begin{aligned}
m(\ddot{\vect p}-\vect g)&=\mat R(\vect F_A+\vect F_T),\\
\mat I\dot{\vect\omega}+\vect\omega\times\mat I\vect\omega
&=\vect M_A+\vect M_T,
\end{aligned}\label{eq:dynamics}
\end{equation}
where $m$ is mass, $\vect g$ is gravity, and subscripts $A$ and $T$ denote aerodynamic and propulsor loads in body coordinates. The aerodynamic moment includes rate damping. Wind-tunnel-based periodic coefficient fits describe aerodynamic loading over the full angle-of-attack range, so the model covers the large attitude excursions involved in the maneuvers.

The propulsors produce no body-lateral force. With a common longitudinal arm $l_x$, the total normal thrust $F_{T,z}$ and propulsor pitching moment $M_{T,y}$ satisfy
\begin{equation}
F_{T,y}=0,\qquad M_{T,y}=-l_xF_{T,z}.
\label{eq:coupling}
\end{equation}
Thus, the normal thrust needed for pitch control also changes the translational force balance. This relation determines how the attitude can evolve along a prescribed position trajectory. At large aerodynamic angles, its effect becomes appreciable as aerodynamic support and the required pitching moment change together.

A maneuver is specified by its characteristic geometry and ordered attitude features. Position and heading define the path-following portion, while roll sequences can additionally prescribe rotations about the longitudinal axis. Trajectory design adjusts the remaining motion under the aircraft dynamics and actuator bounds. The normalized rotor-speed commands lie in $[0,1]$, and the vectoring angles are limited to $\pm0.6$~rad. These requirements guide the choice of path scale, curvature and timing.

\begin{figure}[t]
\centering
\includegraphics[width=\linewidth]{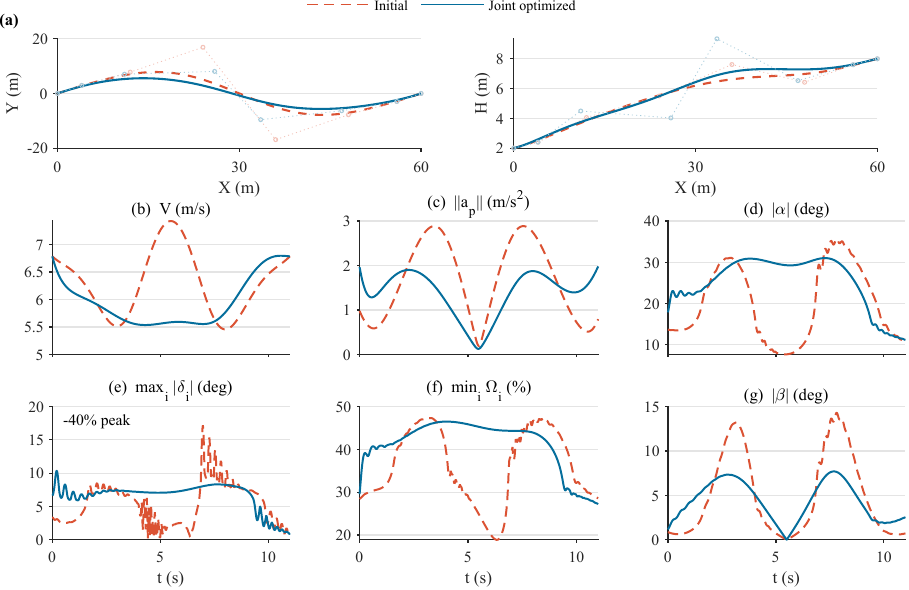}
\caption{S-shaped climb before and after joint geometry and timing optimization. The endpoint motion and duration are fixed.}
\label{fig:optimization}
\end{figure}

\section{Trajectory Generation and Tracking}\label{sec:method}
The force and moment relations above provide the basis for recovering an attitude history along a path. Geometry and timing can then be adjusted according to the resulting aerodynamic and propulsor demands. The recovered attitude and propulsor demands connect the trajectory design variables to the aircraft dynamics.

\subsection{Attitude Recovery from Force and Moment Balance}

The planning output is $\vect\sigma=[\vect p^\top,\chi]^\top$, where $\chi$ specifies the azimuth of the body longitudinal axis. Its inclination $\vartheta$ and the rotation $\mu$ about that axis determine the remaining attitude freedom. A convenient body basis is
\begin{equation}
\begin{aligned}
\vect x_b&=[\cos\vartheta\cos\chi,\cos\vartheta\sin\chi,-\sin\vartheta]^\top,\\
\vect y_\chi&=[-\sin\chi,\cos\chi,0]^\top,\\
\vect z_\chi&=(\vect x_b\times\vect y_\chi)/\|\vect x_b\times\vect y_\chi\|,\\
\vect y_b&=\cos\mu\,\vect y_\chi+\sin\mu\,\vect z_\chi,\\
\vect z_b&=\vect x_b\times\vect y_b,\qquad
\mat R=[\vect x_b,\vect y_b,\vect z_b].
\end{aligned}\label{eq:basis}
\end{equation}
The rotation matrix gives angular velocity through $[\vect\omega]_\times=\mat R^\top\dot{\mat R}$, where $[\cdot]_\times$ is the skew-symmetric cross-product matrix. Differentiation gives angular acceleration. The loads required by a candidate attitude history follow as
\begin{equation}
\begin{aligned}
\vect F_T^{\mathrm{req}}&=\mat R^\top m(\ddot{\vect p}-\vect g)-\vect F_A,\\
\vect M_T^{\mathrm{req}}&=\mat I\dot{\vect\omega}+\vect\omega\times\mat I\vect\omega-\vect M_A.
\end{aligned}\label{eq:required}
\end{equation}
Substituting the required loads into the propulsor relations gives the two compatibility equations
\begin{equation}
F_{T,y}^{\mathrm{req}}=0,\qquad
F_{T,z}^{\mathrm{req}}+\frac{M_{T,y}^{\mathrm{req}}}{l_x}=0.
\label{eq:compatibility}
\end{equation}
The first condition determines the lateral balance, and the second enforces the coupling in Eq.~\eqref{eq:coupling}.

The required pitching moment contains the angular acceleration generated by $\vartheta$ and $\mu$. After solving the lateral-force equation for $\mu$, the second condition in Eq.~\eqref{eq:compatibility} gives an internal attitude equation of the form
\begin{equation}
A(\vartheta,\vect\sigma^{(4)})\ddot\vartheta+
B(\vartheta,\dot\vartheta,\vect\sigma^{(4)})=0,
\label{eq:internal}
\end{equation}
where $\vect\sigma^{(4)}$ denotes the prescribed output and its derivatives through fourth order. The coefficients collect the inertia, aerodynamic loads and thrust moment arm. This equation retains the entry inclination and inclination rate, so a position--heading history alone does not algebraically determine the attitude. For open maneuvers, boundary conditions are derived from the constant-velocity entry and exit motions. Closed maneuvers use periodic attitude and attitude-rate conditions.

The numerical inversion represents $\vartheta$ and $\mu$ at trajectory samples, differentiates the resulting rotation sequence, and solves the compatibility residuals together with the boundary equations. This yields the attitude history and the required propulsor loads. With lateral arm $l_y$, individual force components are allocated as
\begin{equation}
\begin{aligned}
F_{1x,2x}&=\tfrac12(F_{T,x}^{\mathrm{req}}\pm M_{T,z}^{\mathrm{req}}/l_y),\\
F_{1z,2z}&=\tfrac12(F_{T,z}^{\mathrm{req}}\mp M_{T,x}^{\mathrm{req}}/l_y).
\end{aligned}\label{eq:allocation}
\end{equation}
The upper and lower signs denote propulsors 1 and 2. For the positive axial-thrust references considered here, the commands are
\begin{equation}
\begin{aligned}
T_i&=\sqrt{F_{ix}^2+F_{iz}^2},\qquad
\delta_i=\tan^{-1}(-F_{iz}/F_{ix}),\\
\Omega_i&=\sqrt{T_i/k_T},
\end{aligned}\label{eq:commands}
\end{equation}
where $T_i$ is thrust magnitude and $k_T$ is the thrust coefficient for normalized rotor speed. The attitude matrix is converted to a scalar-first unit quaternion for tracking.

Prescribed roll motions require additional attitude constraints. In these portions, the given geometric and attitude features are incorporated into trajectory construction while enforcing the same aircraft dynamics. The cloverleaf and half-Cuban examples below recover internal pitch motion along their prescribed maneuver geometry. The straight rolling sequence also prescribes the longitudinal-axis rotation. These different maneuver requirements share the force and moment relations in Eqs.~\eqref{eq:dynamics} and~\eqref{eq:coupling}.

\subsection{Joint Geometry and Timing Optimization}
A quintic B-spline represents the position curve, and a second B-spline sets its progress rate. Let $s\in[0,1]$ denote path progress and $r=t/T$ normalized time, with maneuver duration $T$. The two representations are
\begin{equation}
\begin{aligned}
\vect p(s)&=\sum_i N_{i,5}(s)\vect P_i,\\
\zeta(r)&=\frac{ds}{dr}=\sum_j B_j(r)\zeta_j>0,\qquad
\dot s=\zeta/T,
\end{aligned}\label{eq:splines}
\end{equation}
where $N_{i,5}$ and $B_j$ are spline basis functions, $\vect P_i$ are position control points and $\zeta_j$ are progress-rate control values. Repeated endpoint knots impose the endpoint positions, and adjacent control points set the endpoint velocities. The timing constraints $\zeta(0)=\zeta(1)=1$ and $\int_0^1\zeta(r)dr=1$ preserve the endpoint velocities and total duration while allowing nonuniform progress. The corresponding velocity and acceleration are
\begin{equation}
\begin{aligned}
\dot{\vect p}&=\vect p_s\dot s,\\
\ddot{\vect p}&=\vect p_{ss}\dot s^2+\vect p_s\ddot s,
\end{aligned}\label{eq:timing}
\end{equation}
where subscripts denote differentiation with respect to $s$. Changing the curve alters its curvature, and changing progress redistributes acceleration along it.

Let $\vect\theta$ collect the free position and timing control values. The optimization has the form
\begin{equation}
\begin{aligned}
\min_{\vect\theta}\quad &
w_a(a_{p,\max}/a_{p,s})^2+w_D D^2\\
&+w_\delta(\dot\delta_{\max}/\dot\delta_s)^2+J_{\mathrm{smooth}}\\
\mathrm{s.t.}\quad &\vect h(\vect\theta)=\vect0,\qquad
\vect c(\vect\theta)\leq\vect0.
\end{aligned}\label{eq:optimization}
\end{equation}
Here $a_{p,\max}$ is peak path acceleration, $\dot\delta_{\max}$ is maximum absolute vectoring rate, and $a_{p,s}$ and $\dot\delta_s$ are normalization scales. The nonnegative weights are $w_a,w_D,w_\delta$, and $J_{\mathrm{smooth}}$ penalizes second differences of the position control points and departure from uniform progress. The normalized demand $D$ is the largest ratio of peak angle of attack, sideslip, vectoring angle and rotor speed to their comparison scales. Endpoint conditions and maneuver constraints retain the intended motion, such as the order of loops, straight rolling segments and recovery directions. Trajectory inversion supplies the aerodynamic angles and actuator histories used to evaluate each candidate. An alternating coordinate search updates timing and geometry with progressively smaller steps. Candidate selection first checks the maneuver and physical constraints, and then compares the objective. The selected design is re-evaluated on a finer trajectory grid to obtain the tracking reference.

The normal acceleration on a fixed curve is $a_n=v_\ell^2\kappa$, where $v_\ell$ is arc-length speed and $\kappa$ is curvature. Slowing a curved segment reduces its normal acceleration, but also reduces dynamic pressure and aerodynamic support. For a prescribed duration, that change redistributes speed elsewhere along the path. Geometry optimization can additionally reduce curvature. Joint adjustment therefore addresses the competing demands for acceleration, aerodynamic support and thrust vectoring.

Algorithm~\ref{alg:generation} summarizes the generation procedure. It takes the maneuver geometry and boundary requirements as inputs and returns the state and propulsor references used by the controller.

\begin{algorithm}[t]
\caption{Geometry and timing design with trajectory inversion}
\label{alg:generation}
\begin{algorithmic}[1]
\REQUIRE Initial geometry, heading, maneuver features, boundary conditions and physical bounds
\STATE Initialize the position control points and positive timing law.
\REPEAT
\STATE Evaluate position derivatives and initialize the attitude history.
\STATE Solve Eq.~\eqref{eq:compatibility} with the open or periodic boundary conditions.
\STATE Recover angular motion and allocate the individual propulsor commands using Eqs.~\eqref{eq:allocation} and~\eqref{eq:commands}.
\STATE Evaluate the objective and constraints in Eq.~\eqref{eq:optimization}.
\STATE Update timing and geometry, retaining feasible improvements.
\UNTIL{the coordinate search has converged}
\STATE Re-evaluate the selected trajectory on a finer grid.
\RETURN Position, velocity, attitude, angular-rate and propulsor-command references
\end{algorithmic}
\end{algorithm}

\subsection{Closed-Loop Evaluation}

The generated position, velocity, quaternion and angular-rate references are tracked with an NMPC--INDI controller. NMPC predicts the rigid-body motion and penalizes state tracking errors and control changes. Quaternion error is formed from the relative rotation between the predicted and reference attitudes. INDI converts the acceleration command into updates of the four propulsor inputs using the local force and moment effectiveness.

The position and velocity errors are $\vect e_p=\vect p-\vect p^{\mathrm{ref}}$ and $\vect e_v=\vect v-\vect v^{\mathrm{ref}}$. Attitude tracking uses the principal angle of the relative unit quaternion,
\begin{equation}
\gamma_R=2\cos^{-1}\!\left(\left|\vect q_a^\top\vect q_a^{\mathrm{ref}}\right|\right),
\end{equation}
where $\vect q_a$ and $\vect q_a^{\mathrm{ref}}$ are the actual and reference scalar-first quaternions. The body-rate error is $\vect e_\omega=\vect\omega-\vect\omega^{\mathrm{ref}}$. These quantities penalize departures from both the intended path and attitude motion over the prediction horizon. The predicted axial and angular accelerations provide the command $\vect z_c=[a_x^b,\dot{\vect\omega}^{\top}]^\top$ for INDI, where $a_x^b$ is inertial acceleration resolved along the body longitudinal axis.

At the current aircraft state, the controlled acceleration is affine in the four propulsor force components. Solving this relation for the commanded acceleration and its measured estimate gives the force vectors $\vect f_c$ and $\vect f_m$. If $\mathcal A$ maps force components into actuator commands using Eq.~\eqref{eq:commands}, the incremental update is
\begin{equation}
\vect u^+=\vect u_{\mathrm{app}}+
\mathcal A(\vect f_c)-\mathcal A(\vect f_m),
\end{equation}
where $\vect u_{\mathrm{app}}$ is the currently applied input. Vectoring and rotor-speed bounds are applied before the updated commands enter the aircraft dynamics.

The simulations integrate the aircraft dynamics under the applied actuator commands. The controller uses model-derived acceleration estimates, and the plant and controller share the modeled aircraft parameters. The plotted paths and aircraft outlines use the resulting closed-loop position and attitude. Thus, the trajectory comparisons show the response produced by the tracking controller, together with the original maneuver geometry.

\section{Representative Results and Discussion}\label{sec:results}
The results first examine how geometry and timing affect demand, and then illustrate three aerobatic sequences. The selected maneuvers cover a spatial cloverleaf, a half-Cuban combination and straight rolls. Each sequence includes its closing or recovery segment. All reported tracking errors are evaluated against the generated continuous reference.

\subsection{Effect of Geometry and Timing}
Figure~\ref{fig:optimization} compares the initial and jointly optimized S-shaped climbs with the same endpoint positions, velocities and duration. Timing optimization on the initial geometry reduces thrust-vectoring demand, while peak path acceleration changes from $0.29g$ to $0.31g$. Joint adjustment reduces peak acceleration to $0.20g$, and the maximum vectoring angle and rate decrease from $17.2^\circ$ and $171^\circ/\mathrm{s}$ to $10.4^\circ$ and $46^\circ/\mathrm{s}$. The optimized path retains the two opposite lateral turns.

The different acceleration responses follow from Eq.~\eqref{eq:timing}. A time law can move acceleration between segments of the initial curve, whereas geometry changes also alter the curvature that generates normal acceleration. In this comparison, adjusting both reduces acceleration and vectoring demand together. The maneuver constraints preserve its two-turn structure as the path changes.

\subsection{Looping and Rolling Sequences}
\begin{figure}[t]
\centering
\includegraphics[width=\linewidth]{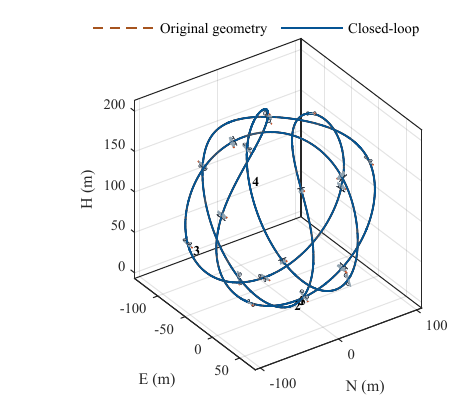}
\caption{Cloverleaf with four quarter rolls (F1). Dashed curves show the original geometry, and solid curves show the closed-loop path. Enlarged aircraft use actual attitudes and applied vectoring commands at equal arc-length intervals. Numbers indicate flight order.}
\label{fig:looping}
\end{figure}
\begin{figure}[t]
\centering
\includegraphics[width=\linewidth]{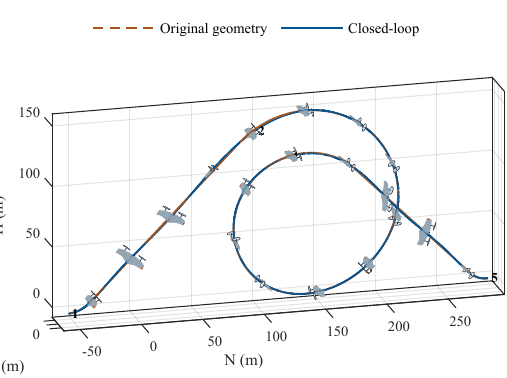}
\caption{Reverse half-Cuban and half-Cuban sequence (F4), using the same conventions as Fig.~\ref{fig:looping}.}
\label{fig:cuban}
\end{figure}

Figures~\ref{fig:looping} and~\ref{fig:cuban} show the cloverleaf with four quarter rolls and the reverse half-Cuban/half-Cuban combination. Dashed orange curves denote the original maneuver geometry, and solid blue curves show the actual closed-loop trajectory. Aircraft outlines use the actual attitude and applied vectoring commands at equal arc-length intervals. The enlarged outlines and numbered markers indicate attitude evolution and flight order.

The cloverleaf retains its four lobes and quarter-roll order. Its repeated changes in maneuver plane are accompanied by the prescribed body-axis rotations, while the closed-loop trajectory remains close to the intended geometry. The half-Cuban sequence retains the ordered climbing, inversion and recovery portions. Compared with the cloverleaf, its tighter changes in flight direction require a larger specific-force load, as reflected in Table~\ref{tab:tracking}.

\begin{figure}[t]
\centering
\includegraphics[width=\linewidth]{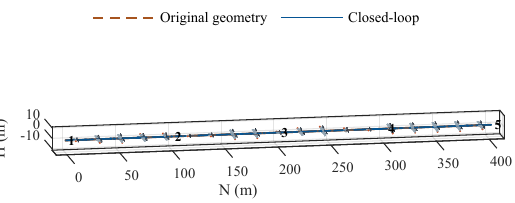}
\caption{Straight rolling sequence (F8): continuous roll, four-point roll and knife-edge hold.}
\label{fig:rolling}
\end{figure}

The straight rolling sequence in Fig.~\ref{fig:rolling} maintains its linear flight path through a continuous roll, a four-point roll and a knife-edge hold. This example separates the geometric path from the required body-axis rotation. The force and moment balance accommodates the attitude changes while preserving the straight segment. Its maximum specific-force load remains close to unity, although the aircraft undergoes substantial attitude motion. Path acceleration alone therefore gives an incomplete description of an aerobatic reference, and the prescribed attitude features are needed to identify the maneuver.

\begin{table}[t]
\centering
\caption{Closed-loop response of the illustrated sequences.}
\label{tab:tracking}
\begin{tabular}{lrrrr}
\toprule
Case & $n_{\max}$ & $\Omega_{\max}$ & $e_{p,\max}$ (mm) & $\gamma_{R,\max}$ (deg)\\
\midrule
F1 & 1.34 & 0.72 & 1.3 & 0.041\\
F4 & 2.94 & 0.70 & 6.7 & 0.77\\
F8 & 1.01 & 0.58 & 4.0 & 0.90\\
\bottomrule
\end{tabular}
\end{table}

Table~\ref{tab:tracking} reports peak specific-force load $n=\|\ddot{\vect p}-\vect g\|/g$, normalized rotor speed $\Omega$, position error $e_p$, and the principal relative-quaternion angle $\gamma_R$. The three sequences remain below a peak load of $3g$. Small position and attitude errors show that the modeled aircraft follows both the generated path and its attitude motion under the stated simulation conditions. Together, the examples demonstrate the use of the force--moment relations for maneuvers with different geometric and attitude features.

\subsection{State Tracking and Applied Inputs}
\begin{figure}[!htbp]
\centering
\includegraphics[width=\linewidth]{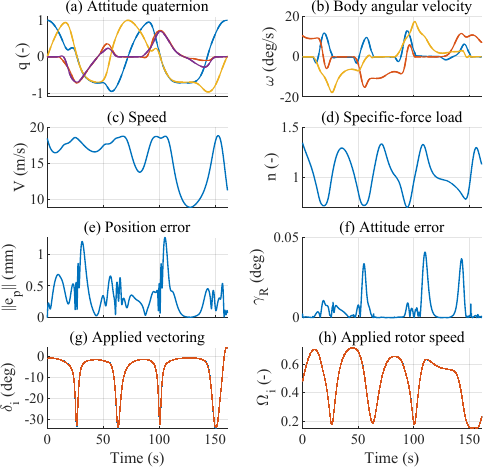}
\caption{F1 closed-loop response. In (a,b), solid curves are actual states and dashed curves are references. Quaternion components $(w,x,y,z)$ use blue, orange, yellow and purple; body rates $(x,y,z)$ use blue, orange and yellow. Applied input channels 1 and 2 use blue and orange.}
\label{fig:f1hist}
\end{figure}

\begin{figure}[!htbp]
\centering
\includegraphics[width=\linewidth]{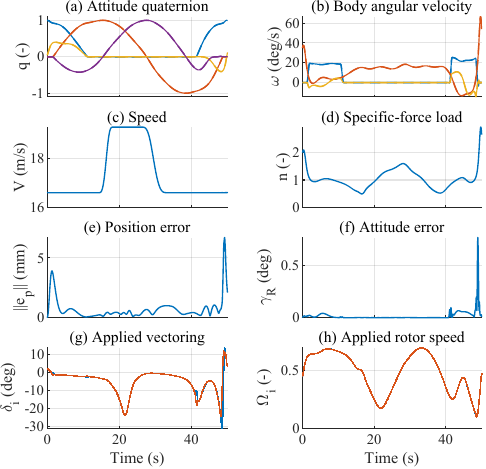}
\caption{F4 closed-loop response. In (a,b), solid curves are actual states and dashed curves are references. Quaternion components $(w,x,y,z)$ use blue, orange, yellow and purple; body rates $(x,y,z)$ use blue, orange and yellow. Applied input channels 1 and 2 use blue and orange.}
\label{fig:f4hist}
\end{figure}

\begin{figure}[!htbp]
\centering
\includegraphics[width=\linewidth]{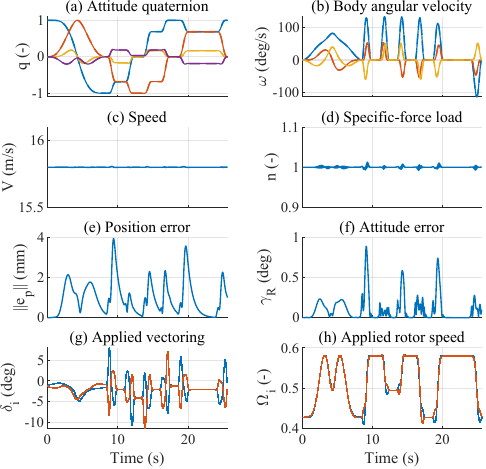}
\caption{F8 closed-loop response. In (a,b), solid curves are actual states and dashed curves are references. Quaternion components $(w,x,y,z)$ use blue, orange, yellow and purple; body rates $(x,y,z)$ use blue, orange and yellow. Applied input channels 1 and 2 use blue and orange.}
\label{fig:f8hist}
\end{figure}

Figures~\ref{fig:f1hist}--\ref{fig:f8hist} show the closed-loop histories corresponding to the three trajectories. Quaternion and body-rate panels compare the generated references with the actual states. The remaining panels show speed, specific-force load, tracking errors and applied propulsor commands. Position and attitude errors use common saved reference instants. The applied commands are plotted with their holding intervals, matching the inputs used in the state integration.

The cloverleaf follows the repeated changes in attitude with small tracking errors over the four lobes. The half-Cuban combination produces stronger body-rate and load variations during the climbing and inverted recovery portions. The straight rolling sequence retains a nearly constant speed and load while its quaternion components change throughout the roll and knife-edge portions. Together with the position errors, these histories show how the controller tracks attitude changes while preserving the corresponding flight path.

\FloatBarrier
\section{Conclusion}\label{sec:conclusion}
A rudderless dual vectored-thrust flying wing combines VTOL and efficient forward flight, while its coupled translation and pitch actuation complicates aerobatic trajectory generation. This study recovers internal attitude motion through force and moment compatibility and uses the resulting demands to guide joint geometry and timing optimization. The S-shaped climb comparison shows that adjusting geometry and timing together can reduce path acceleration and thrust-vectoring demand while preserving maneuver geometry. Closed-loop simulations of cloverleaf, half-Cuban and straight rolling sequences demonstrate coordinated path and attitude tracking. The method provides a basis for designing aerobatic maneuvers for rudderless vectored-thrust aircraft.

\section*{Data Availability}
Trajectory references and simulated closed-loop responses are available at \url{https://github.com/dynamics-pilot/vector-thrust-maneuver-data}.

\newpage
\bibliographystyle{IEEEtran}
\bibliography{references}

@article{Di2025TaillessVTOL,
  author  = {Di, Weicheng and Wei, Zixing and Liu, Qiqi and Zhang, Jiahui and Xiang, Jinwu and Li, Daochun and Tu, Zhan},
  title   = {Dynamics and Control of Tailless Vectored-Thrust {VTOL} in Hover-to-Forward Transition},
  journal = {Aerospace Science and Technology},
  volume  = {164},
  pages   = {110385},
  year    = {2025},
  doi     = {10.1016/j.ast.2025.110385}
}

@incollection{richter2016polynomial,
  author = {Richter, Charles and Bry, Adam and Roy, Nicholas},
  title = {Polynomial Trajectory Planning for Aggressive Quadrotor Flight in Dense Indoor Environments},
  booktitle = {Robotics Research: The 16th International Symposium {ISRR}},
  editor = {Inaba, Masayuki and Corke, Peter},
  series = {Springer Tracts in Advanced Robotics},
  volume = {114},
  pages = {649--666},
  publisher = {Springer International Publishing},
  address = {Cham, Switzerland},
  year = {2016},
  doi = {10.1007/978-3-319-28872-7_37}
}

@article{foehn2021timeoptimal,
  author = {Foehn, Philipp and Romero, Angel and Scaramuzza, Davide},
  title = {Time-Optimal Planning for Quadrotor Waypoint Flight},
  journal = {Science Robotics},
  year = {2021},
  volume = {6},
  number = {56},
  pages = {eabh1221},
  doi = {10.1126/scirobotics.abh1221}
}

@inproceedings{usenko2017bspline,
  author = {Usenko, Vladyslav and von Stumberg, Lukas and Pangercic, Andrej and Cremers, Daniel},
  title = {Real-Time Trajectory Replanning for {MAVs} Using Uniform {B}-Splines and a 3D Circular Buffer},
  booktitle = {2017 IEEE/RSJ International Conference on Intelligent Robots and Systems ({IROS})},
  publisher = {IEEE},
  address = {Vancouver, BC, Canada},
  month = {24--28 September},
  year = {2017},
  pages = {215--222},
  doi = {10.1109/IROS.2017.8202160}
}

@article{ducard2021review,
  author = {Ducard, Guillaume J. J. and Allenspach, Mike},
  title = {Review of Designs and Flight Control Techniques of Hybrid and Convertible {VTOL} {UAVs}},
  journal = {Aerospace Science and Technology},
  year = {2021},
  volume = {118},
  pages = {107035},
  doi = {10.1016/j.ast.2021.107035}
}

@article{tal2022aerobatic,
  author = {Tal, Ezra and Ryou, Gilhyun and Karaman, Sertac},
  title = {Aerobatic Trajectory Generation for a {VTOL} Fixed-Wing Aircraft Using Differential Flatness},
  journal = {IEEE Transactions on Robotics},
  year = {2023},
  volume = {39},
  number = {6},
  pages = {4805--4819},
  doi = {10.1109/TRO.2023.3301312}
}

@article{lu2025autonomous,
  author = {Lu, Guozheng and Ren, Yunfan and Zhu, Fangcheng and Li, Haotian and Xue, Ruize and Cai, Yixi and Lyu, Ximin and Zhang, Fu},
  title = {Autonomous Tail-Sitter Flights in Unknown Environments},
  journal = {IEEE Transactions on Robotics},
  year = {2025},
  volume = {41},
  pages = {1098--1117},
  doi = {10.1109/TRO.2025.3526102}
}

@article{druecker2023tracking,
  author  = {Dr{\"u}cker, Svenja and Seifried, Robert},
  title   = {Trajectory-Tracking Control from a Multibody System Dynamics Perspective},
  journal = {Multibody System Dynamics},
  volume  = {58},
  pages   = {341--363},
  year    = {2023},
  doi     = {10.1007/s11044-022-09870-9}
}
\end{document}